\documentclass[12pt]{article}

\usepackage[T1]{fontenc}
\usepackage[utf8]{inputenc}
\usepackage{lmodern}           

\usepackage[a4paper, margin=2.5cm]{geometry}
\usepackage{graphicx}
\usepackage{booktabs}
\usepackage{hyperref}
\usepackage{subcaption}
\usepackage[font=small]{caption}
\usepackage{placeins}

\usepackage{amsmath}
\usepackage{amssymb}

\usepackage{makecell}

\usepackage{float}

\usepackage[authoryear]{natbib}
\usepackage{hyperref}

\usepackage{tipa}
\usepackage{tipx}              

\newcommand{\ipa}[1]{\textipa{#1}}

\title{\textbf{Lips-Jaw and Tongue-Jaw Articulatory Tradeoff in DYNARTmo}}

\author{
  Bernd J.~Kröger$^{1,2}$\\[1ex]
  {\small $^{1}$Medical School, RWTH Aachen University, Aachen, Germany}\\
  {\small $^{2}$Kröger Lab, Belgium, \url{www.speechtrainer.eu}}\\[2ex]
}

\date{}

\begin{document}

\clubpenalty=3000
\widowpenalty=3000

\maketitle

\begin{abstract}
This paper investigates how the dynamic articulatory model DYNARTmo accounts for articulatory tradeoffs between primary and 
secondary articulators, with a focus on lips--jaw and tongue--jaw coordination. While DYNARTmo does not implement full task-dynamic 
second-order biomechanics, it adopts first-order task-space gesture specifications comparable to those used in articulatory phonology 
and integrates a simplified mechanism for distributing articulatory effort across multiple articulators. We first outline the 
conceptual relationship between task dynamics and DYNARTmo, emphasizing the distinction between high-level task-space trajectories 
and their low-level articulatory execution. We then present simulation results for a set of CV syllables that illustrate how jaw 
displacement varies as a function of both place of articulation (labial, apical, dorsal) and vowel context (/a/, /i/, /u/). 
The model reproduces empirically attested patterns of articulatory synergy, including jaw-supported apical closures, lower-lip 
elevation in bilabial stops, tongue–jaw co-movement, and saturation effects in labial constrictions. These results demonstrate 
that even with computationally simplified assumptions, DYNARTmo can generate realistic spatio-temporal movement patterns that 
capture key aspects of articulatory tradeoff and synergy across a range of consonant--vowel combinations.
\end{abstract}

\section{Introduction}

Dynamic articulatory modeling has long aimed to capture the complex coordination patterns that 
underlie speech movements. In a recent publication, we introduced \textit{DYNARTmo}, a dynamic 
articulatory model designed primarily for visualization and analysis of speech movement 
patterns \citep{Kroeger2025_DYNARTmo}. DYNARTmo is deliberately simpler than classical task-dynamic 
frameworks such as Saltzman and Munhall’s model of articulatory dynamics \citep{Saltzman1986,
SaltzmanMunhall1989}. Importantly, DYNARTmo does not implement the full task-dynamic architecture 
with second-order dynamical systems and gestural tracking in a formally defined task space.

Nevertheless, DYNARTmo adopts the same mathematical formulation for generating gestures in task space, 
and it employs an approach comparable to that of Articulatory Phonology \citep{BrowmanGoldstein1992} 
for constructing gestural scores. While the spatio-temporal coordination between gestures 
(inter-gesture articulator coordination) is therefore modeled in a conceptually similar manner, 
the spatial tradeoff of gesture-induced articulator displacements---such as those shared between the lips and the lower jaw, or between the 
tongue and the lower jaw (i.e., intra-gesture articulator coordination)---is treated differently. 
In DYNARTmo, these intra-gesture coordination 
patterns are implemented through an alternative mechanism, allowing the model to explicitly 
represent how multiple articulators jointly contribute to a single constriction.

One particularly relevant articulatory tradeoff concerns the division of labor between tongue and 
jaw or lips and jaw during the formation of oral constrictions, for instance in labial closure 
gestures during production of consonants like \ipa{/p/, /b/, /m/} or apical closure gestures 
during production of consonants like \ipa{/t/, /d/, /n/}. Empirical studies have repeatedly 
shown that the jaw systematically contributes to the formation of many consonantal constrictions, 
even when the constriction target is primarily specified for a superior articulator (tongue tip, 
tongue blade, or lips). Whether the constriction is apical or labial, the jaw typically rises in 
concert with the primary articulator—tongue tip or lower lip—--thereby reducing the relative displacement 
needed and enabling the gesture to be achieved with less extreme movement \citep{Mooshammer1995,
Harshman1977,Westbury1988,LofqvistGracco1997}. In the articulatory-phonology framework, such effects 
are interpreted as reflecting the cooperative action of multiple articulators acting as a synergy 
toward a single constriction target.

From a biomechanical and energetic perspective, articulatory tradeoffs between coordinated articulators 
may reflect general principles of economy of effort. \citet{Lindblom1990} and subsequent work on the 
``economy of speech gestures'' have argued that dividing a displacement requirement across multiple 
articulators can reduce total articulatory cost. If a total displacement of magnitude
$l = 2$ is required to achieve a constriction, a single articulator must produce the entire displacement, 
yielding an estimated energetic cost of $l^2 = 4$. If, however, the displacement is shared between two 
cooperating articulators, such that
$l = l_1 + l_2 = 1 + 1$,
the combined cost is $l_1^{2} + l_2^{2} = 1 + 1 = 2$, 
i.e., reduced by half.  This simplified energetic argument illustrates why coordinated tongue--jaw 
synergies, such as apical closures achieved jointly by tongue tip elevation and jaw raising, may represent efficient strategies for achieving articulatory targets.

In the present paper (i) the task-dynamics approach as well as our DYNARTmo approach 
will be introduced and (ii) some examples for tongue--jaw tradeoffs will be introduced for DYNARTmo. We 
will illustrate how the DYNARTmo system naturally produces synergistic articulator movements in a 
range of consonant--vowel contexts and demonstrate context effects that closely resemble empirically 
reported jaw--tongue coordination patterns.

\section{Task-Dynamics and DYNARTmo}

\subsection{Classical Task Dynamics}

The task-dynamic model \citep{Saltzman1979,Saltzman1986,SaltzmanMunhall1989}, developed at Haskins 
Laboratories, treats speech gestures as dynamical units defined in a task space, e.g.\ lip aperture, 
tongue-tip constriction degree, and velic opening. Gestures are implemented as critically damped 
second-order dynamical systems:

\[
\ddot{x} + B \dot{x} + K (x - x_{\mathrm{target}}) = 0,
\]

where \( x \) is a task variable (e.g., lip aperture), \( x_{\mathrm{target}} \) its gestural target, 
and \( B, K \) damping and stiffness parameters. These task variables are transformed to 
model-articulator variables through a non-linear kinematic mapping:

\[
x = f(q),
\]

where \( q \) denotes the articulator state vector (jaw position, tongue body position, tongue tip 
position, lip protrusion, etc.). Linear approximations of these mappings, often expressed through 
Jacobian matrices, connect task-space dynamics to articulator commands:

\[
\dot{q} = J^{-1}(q)\,\dot{x}.
\]

Jordan decomposition and stability analyses are used to ensure well-defined movement trajectories, 
and the model is typically implemented with forward and inverse kinematics linking constriction 
tasks to articulator motions.

\subsection{Task Space vs.\ Model-Articulator Space}

A key distinction in articulatory phonology is the separation between task space and model-articulator
space.  

\paragraph{Task Space:}
Task space defines the linguistically relevant constriction variables. In the case of DYNARTmo, 
the task space is augmented by explicitly separating vocalic and consonantal states. Consonantal 
constriction variables consist of constriction \emph{degree} and \emph{location} along the vocal tract 
(from glottis to lips). Constriction degree traditionally distinguishes full closure from near closure 
in order to separate plosives from fricatives; in DYNARTmo this concept is extended to differentiate 
all major consonantal sound types. Thus, DYNARTmo specifies primary articulator forms for separating 
full closure, a fricative constriction, a lateral constriction (with lateral opening), an approximant 
constriction, a vibrant constriction, and so on \citep{Kroeger2025_DYNARTmo}.

The concepts of constriction degree and location, as introduced in Articulatory Phonology 
\citep{BrowmanGoldstein1992}, can also be applied to vowels. Edge vowels such as /a/ and /i/ can be 
described in terms of pharyngeal or palatal constrictions, while /u/ involves a labial and velar 
constriction. However, for vowels the configuration of the \emph{wide cavities} between constrictions 
is equally important. From the viewpoint of vocal-tract acoustics, it is the entire pattern of narrow 
and wide cavity segments extending from glottis to lips that defines the vocalic state. For this reason, 
DYNARTmo introduces the variable \emph{vocal-tract form} as the defining vocalic task variable.

From a linguistic perspective two other articulators are essential: the velum, which determines the 
degree of velopharyngeal constriction, and the arytenoids, which determine the degree of glottal opening. 
Velopharyngeal control distinguishes nasal from oral sounds, and further separates oral sonorants 
(laterals, approximants) from obstruents (plosives, fricatives).

Thus, for a linguistically complete simulation of speech production, the task space must distinguish at 
least four major categories of gestures. These categories correspond directly to four distinct 
task-space states and are reflected in the structure and ordering of gesture tiers within the gesture 
score (see, e.g., \citealp{Kroeger2025_DYNARTmo_Gestures}).

\begin{enumerate}
    \item vocalic shape-forming gestures,
    \item consonantal constriction-forming gestures,
    \item velopharyngeal constriction-forming gestures,
    \item glottal constriction-forming gestures.
\end{enumerate}

In addition, the pulmonary system must be represented to define (a) a constant subglottal-pressure 
maintenance gesture and (b) inhalation and exhalation gestures, in order to separate pre-utterance, 
post-utterance, and utterance-producing states.

\paragraph{Model-Articulator Space:}
Whereas the task space represents a cognitive--linguistic and premotor (or high-level sensorimotor) 
specification---task-space trajectories can be interpreted as sensorimotor predictions or imaginations, 
comparable to the FACTS approach (\citealp{Parrell2019_FACTS})---the model-articulator space defines 
the primary motor level together with its execution organs, i.e., the model articulators. At the 
model-articulator level, the central problem is how to realize the linguistically defined gestures, or 
sensorimotor expectations, that are already specified in task space as idealized trajectories. This 
requires solving the second-order dynamical equations (spring--damping--mass systems) associated with 
articulatory motion, as well as determining how different articulators (primary and secondary) share 
or trade off their contributions to the realization of a single gesture, and how each articulator 
contributes to the execution of simultaneously overlapping vocalic and consonantal gestures.

Thus, a central task of task dynamics is to compute the mapping between these two spaces---task space 
and model-articulator space---in order to explain (i) how a single constriction task (e.g., alveolar 
closure) can be realized by multiple articulatory configurations, and (ii) how overlapping gestures 
jointly satisfy (a) the physical dynamics of the articulatory system, including forces, damping, and 
mass, and (b) the inherent redundancy that allows for articulatory tradeoffs, such as tongue--jaw or 
lip--jaw synergies during apical or labial constriction formation.

\subsection{First-Order Task Specifications vs.\ Second-Order Articulatory Execution}

As outlined above, it is important to distinguish between the mathematical formulation of gestural 
trajectories in task space and the physical implementation of these trajectories by the articulators. 
In both the task-dynamic framework and in DYNARTmo, the spatio-temporal evolution of a gesture is 
fundamentally specified as a \emph{first-order} trajectory in task space, \citep[e.g.,][]{Kroeger2025_DYNARTmo}. 
Such trajectories represent high-level motor expectations or linguistic–cognitive specifications of how 
a constriction variable (e.g., lip aperture in a labial closing gesture) should change over time.

The \emph{second-order} dynamical system commonly associated with task dynamics enters only at the 
level of articulatory execution. When a task-space trajectory is mapped onto the model-articulator 
space, the resulting movement must be realized by articulators with mass, inertia, and neuromuscular 
control properties. The corresponding second-order dynamics---involving forces, stiffness, damping, 
and articulator coupling---reflect the biomechanical and neuromuscular characteristics of the vocal-tract 
system, rather than the linguistic specification of the gesture itself. This distinction is also 
emphasized in hierarchical approaches to motor control \citep[e.g.,][]{Jordan1985,Jordan1995} 
and in articulatory-phonology accounts separating gestural planning from physical realization 
\citep{BrowmanGoldstein1992,SaltzmanMunhall1989}.

From this perspective, gestural overlap at the task level (e.g., the temporal coincidence of a 
labial closing gesture and a vowel gesture in a /ba/ sequence) does not imply direct interaction 
between the gestures themselves. The interaction emerges primarily at the articulatory-execution 
level, where two simultaneously active task trajectories must be jointly mapped onto a shared set of 
articulators. The resulting movements therefore reflect both the high-level task specifications and 
the dynamic properties of the articulator system, including intra-articulatory coupling and biomechanical 
constraints.

This conceptual separation clarifies why task-space trajectories can be expressed as first-order 
expectations while the articulatory system requires second-order dynamics for physical realization.

\section{Modeling the Lips-Jaw and Tongue-Jaw Articulatory Tradeoff in DYNARTmo}

\subsection{The Concept and Principles}

In DYNARTmo we intentionally introduced several ``hard'' simplifications. From a scientific perspective, 
these simplifications limit the biomechanical realism of the model; however, they significantly reduce 
computational load and thus make the approach attractive for potential real-time implementations or for 
deployment as an application on resource-limited devices.

(i) Primary articulators directly follow the task-space trajectories without an intervening biomechanical 
execution model.  

(ii) For both primary and secondary articulators, no explicit motion-equation solutions are computed.

Both simplifications can be motivated by the observation that, in speech production, the mass and inertial 
properties of the articulators play only a minor role compared to the direct neuromuscular control exerted 
on them. In contrast to limb or arm--hand systems---for which the task-dynamics framework was originally 
developed (e.g., \citealp{Jordan1985,Jordan1995,Hogan1985,FlashHogan1985,Bizzi1991,
ShadmehrMussaIvaldi1994,TodorovJordan2002,WolpertKawato1998})---the 
second-order dynamics associated with mass–spring–damper systems are far less dominant in vocal-tract motion. 
The muscular control system is capable of executing articulatory trajectories that remain very close to the 
task-space trajectories, i.e.\ to the sensorimotor expectations formulated at the planning level.

(iii) Articulator-specific tradeoffs, both for the realization of individual gestures and for the coordination 
of temporally overlapping gestures, can therefore be implemented directly at the level of the articulatory model.
The corresponding principles are the following:

(a) Since vocalic articulation determines the overall vocal-tract shape (see \citealp{Kroeger2025_DYNARTmo}), 
the displacement of the lower jaw can initially be equated with the vocalic lower-jaw displacement.

(b) This principle can only be violated if the vocalic articulation is locally overlapped by consonantal 
constriction or closure formation. In these cases, the displacement of the lower jaw must assume an intermediate 
value between the vocalic lower-jaw displacement and the consonantal lower-jaw displacement. The value of the 
consonantal lower-jaw displacement is determined by how strong the lower-jaw displacement would be for the 
formation of this consonant in isolation (i.e., not in a vocalic context).

(c) Through this averaging of the lower-jaw displacement described in (b), a deformation of the anterior region 
of the tongue dorsum must be modeled, as described in \citealp{Kroeger2025_DYNARTmo}.

In the following sub-section, using simulations of CV combinations (CV syllables), we illustrate how these 
principles affect the vertical displacement of the lower jaw across different consonant--vowel combinations.

\subsection{Simulation Examples}

\begin{figure}[!htbp]
    \centering
    \includegraphics[width=0.35\textwidth]{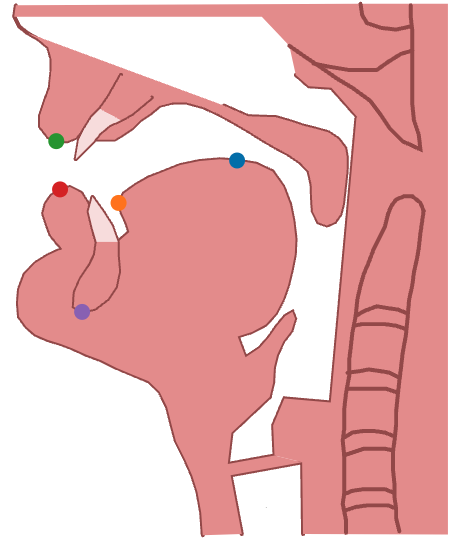}
    \caption{DYNARTmo flesh point locations for upper and lower lip (green, red), tongue tip (orange), 
    tongue dorsum (blue), and the lower jaw (purple) displayed in midsagittal view. Colors of flesh points 
    are identical with colors of flesh point trajectories displayed in Fig. 2 and Fig. 3.}
    \label{fig:fleshpoints_midsagittal}
\end{figure}

\begin{figure}[!htbp]
    \centering

    \begin{subfigure}{0.9\textwidth}
        \centering
        \includegraphics[width=\textwidth]{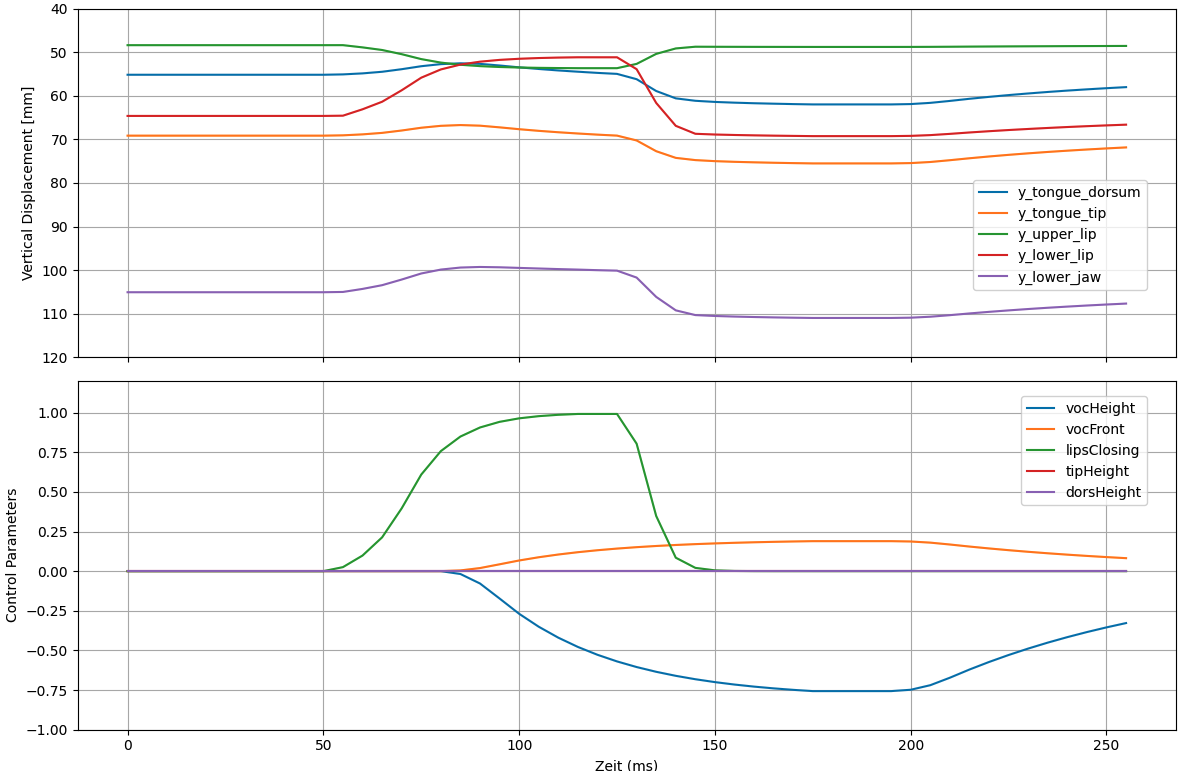}
        \caption{Flesh point and control parameter trajectories for /pa/.}
    \end{subfigure}

    \vspace{0.5cm}

    \begin{subfigure}{0.9\textwidth}
        \centering
        \includegraphics[width=\textwidth]{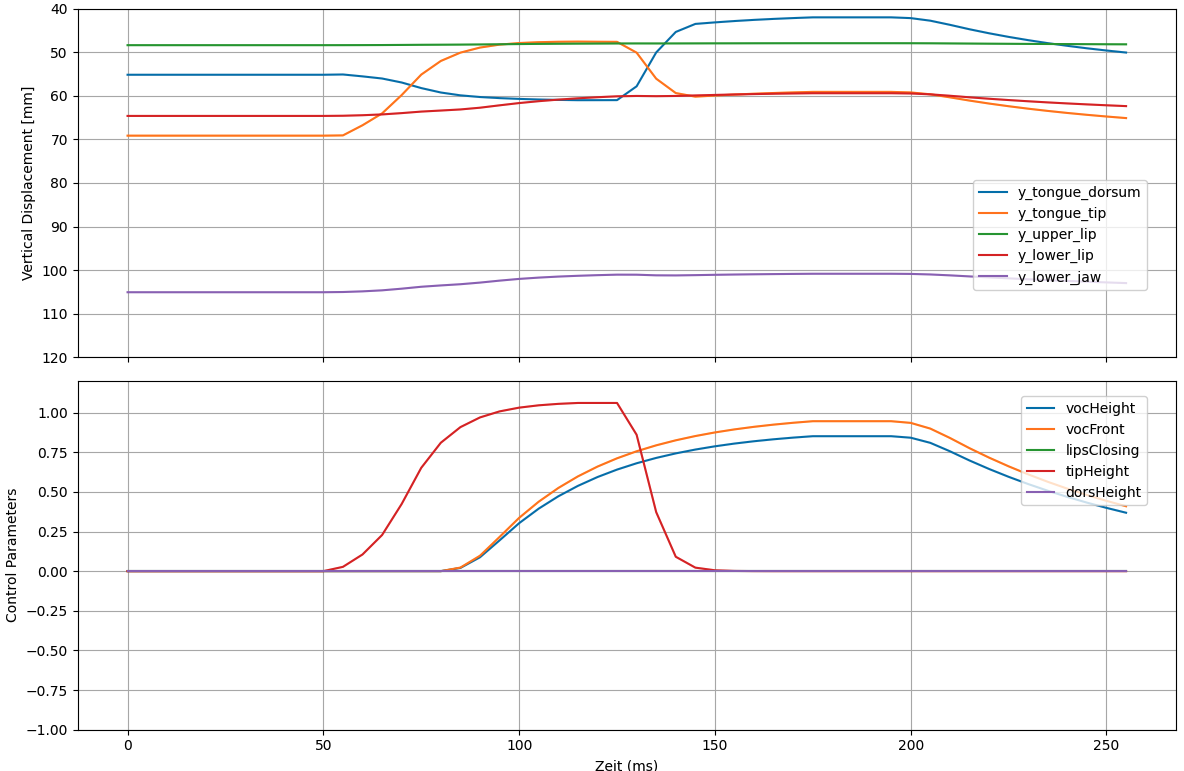}
        \caption{Flesh point and control parameter trajectories for /ti/.}
    \end{subfigure}

    \caption{Flesh point trajectories (vertical displacement) and control parameter trajectories (reflecting 
    task space sensorimotor expectations) generated by DYNARTmo for two syllables; top: /pa/; bottom: /ti/.}
    \label{fig:fleshpoint_pa_ti}
\end{figure}

\begin{figure}[!htbp]
\centering
\begin{subfigure}{0.32\textwidth}
  \centering
  \includegraphics[width=\linewidth]{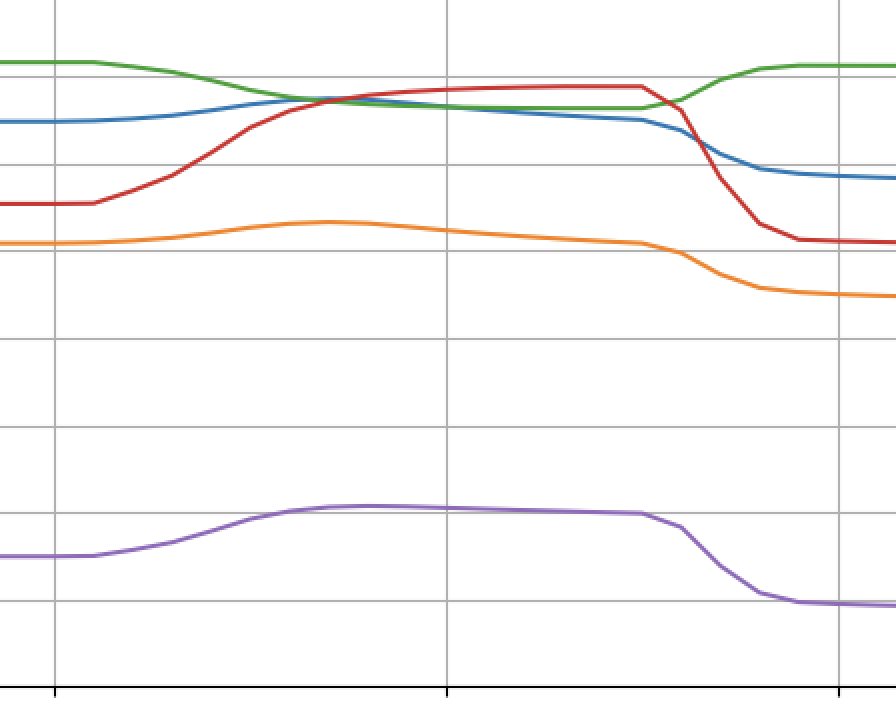}
  \caption{/pa/}
\end{subfigure}
\hfill
\begin{subfigure}{0.32\textwidth}
  \centering
  \includegraphics[width=\linewidth]{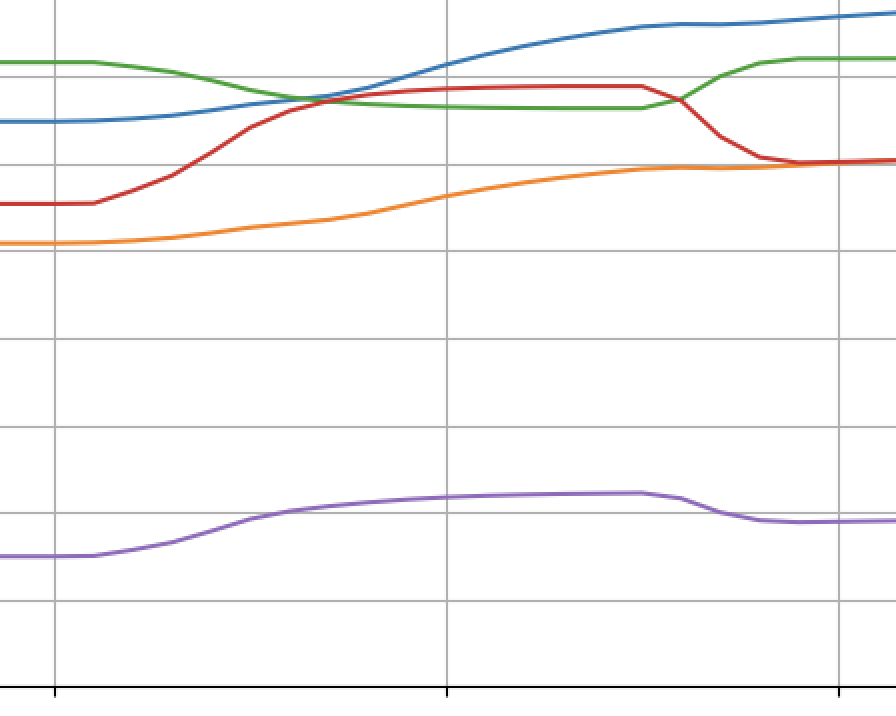}
  \caption{/pi/}
\end{subfigure}
\hfill
\begin{subfigure}{0.32\textwidth}
  \centering
  \includegraphics[width=\linewidth]{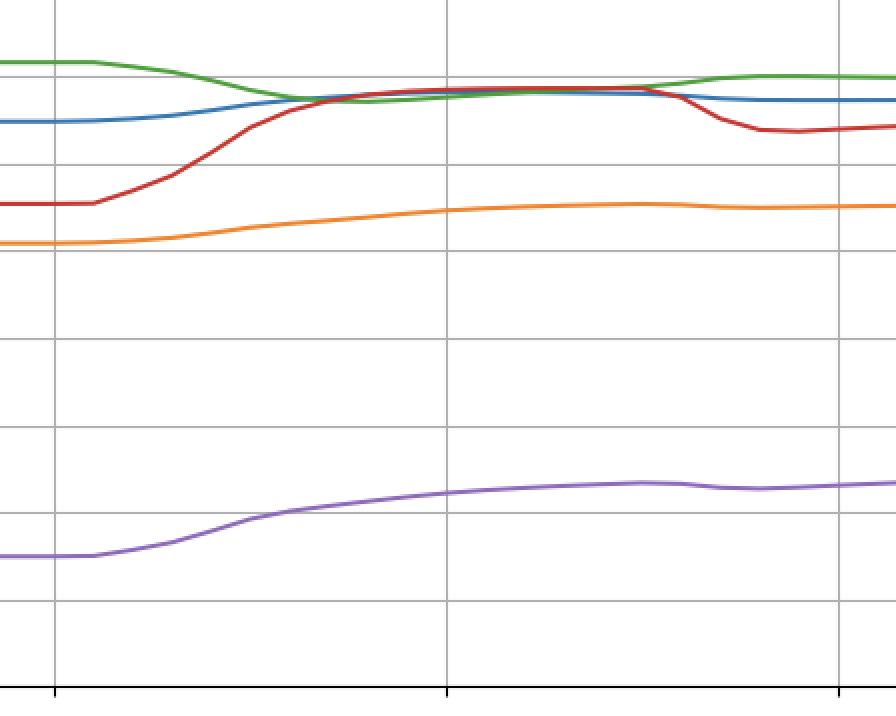}
  \caption{/pu/}
\end{subfigure}
\vspace{1ex}
\begin{subfigure}{0.32\textwidth}
  \centering
  \includegraphics[width=\linewidth]{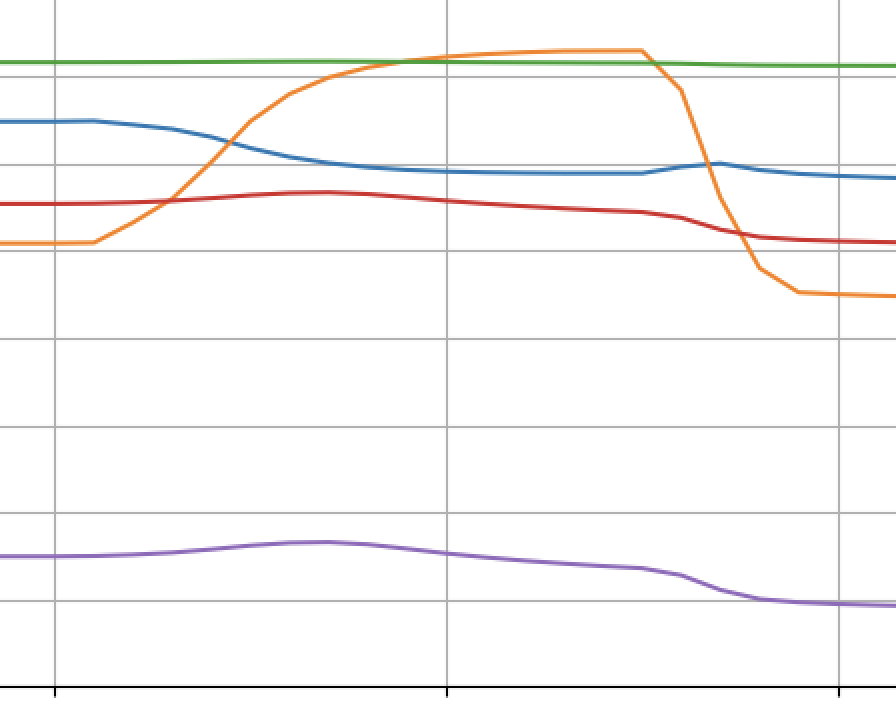}
  \caption{/ta/}
\end{subfigure}
\hfill
\begin{subfigure}{0.32\textwidth}
  \centering
  \includegraphics[width=\linewidth]{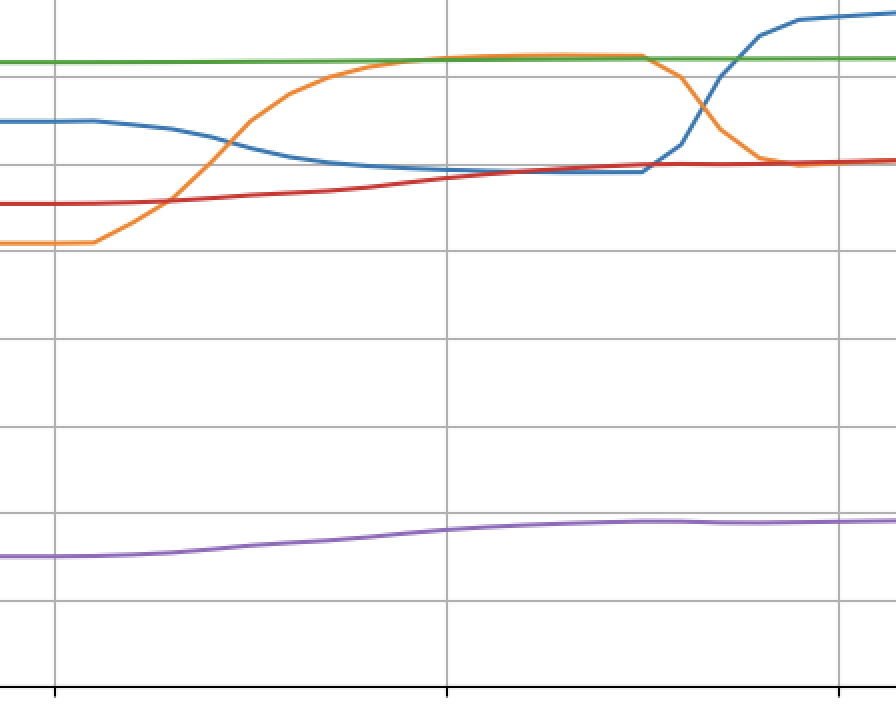}
  \caption{/ti/}
\end{subfigure}
\hfill
\begin{subfigure}{0.32\textwidth}
  \centering
  \includegraphics[width=\linewidth]{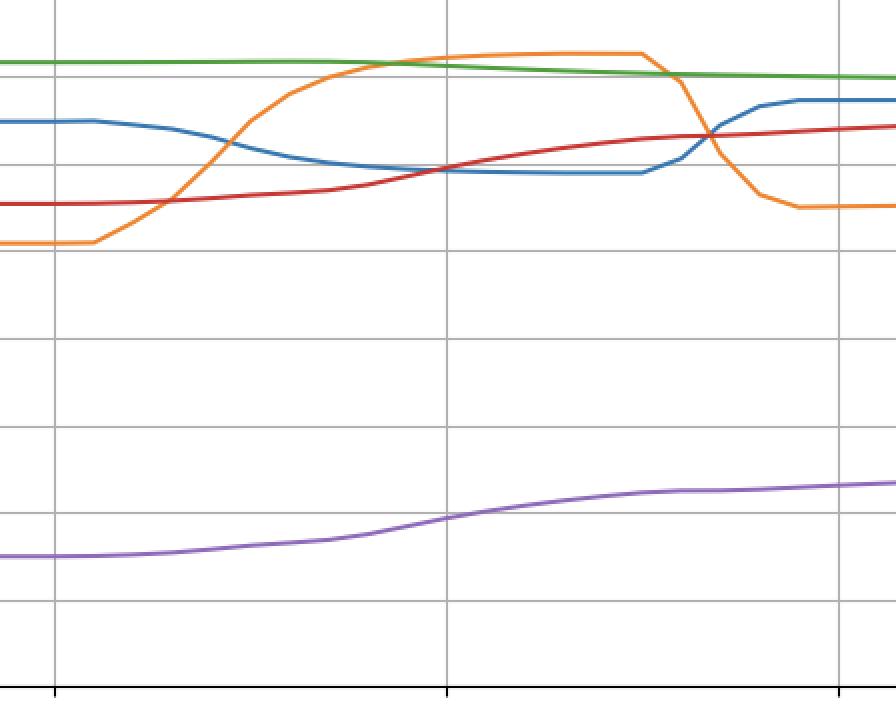}
  \caption{/tu/}
\end{subfigure}
\vspace{1ex}
\begin{subfigure}{0.32\textwidth}
  \centering
  \includegraphics[width=\linewidth]{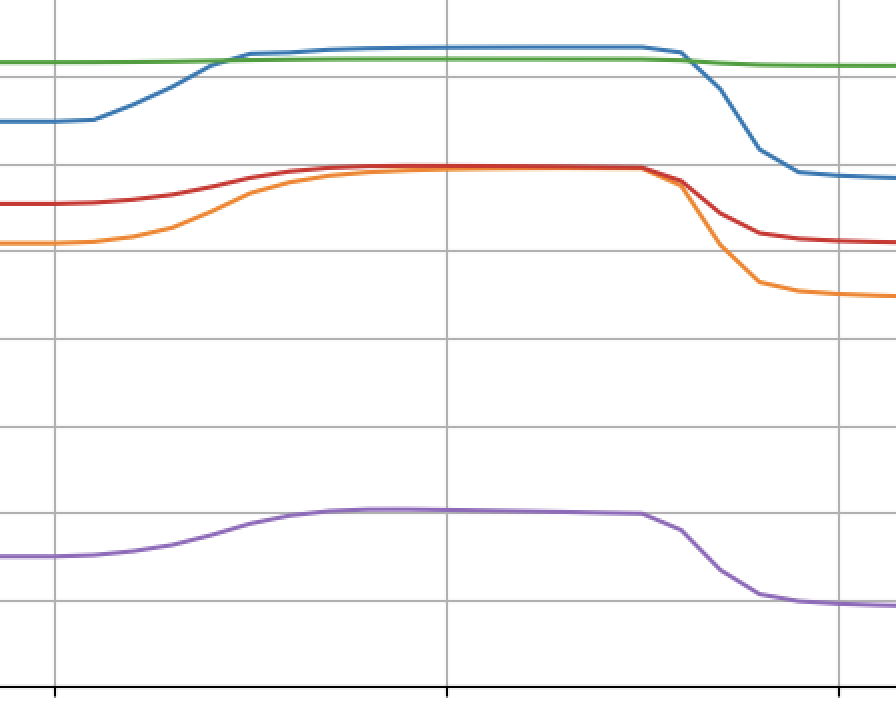}
  \caption{/ka/}
\end{subfigure}
\hfill
\begin{subfigure}{0.32\textwidth}
  \centering
  \includegraphics[width=\linewidth]{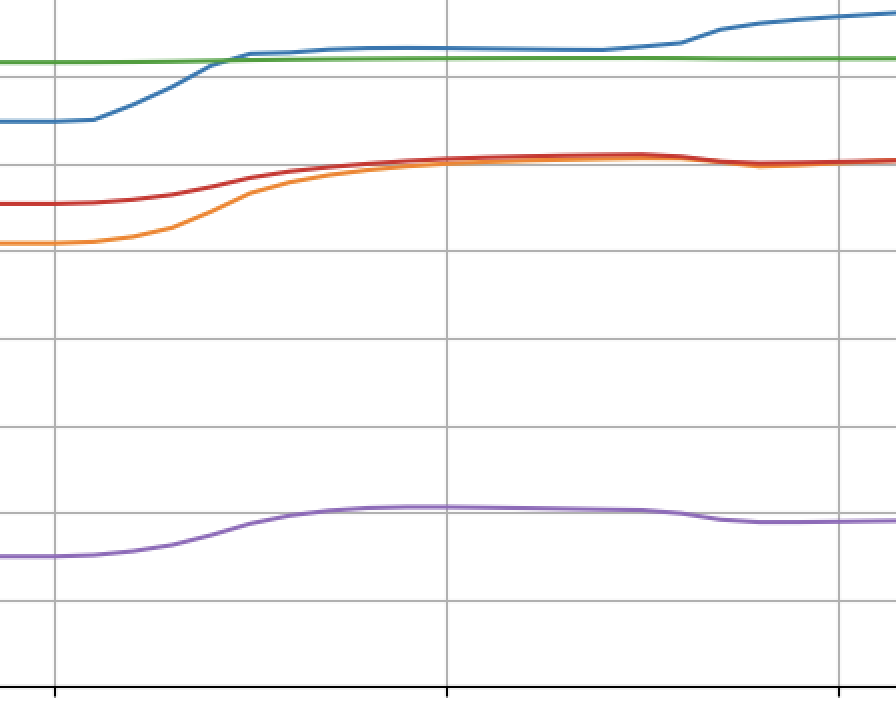}
  \caption{/ki/}
\end{subfigure}
\hfill
\begin{subfigure}{0.32\textwidth}
  \centering
  \includegraphics[width=\linewidth]{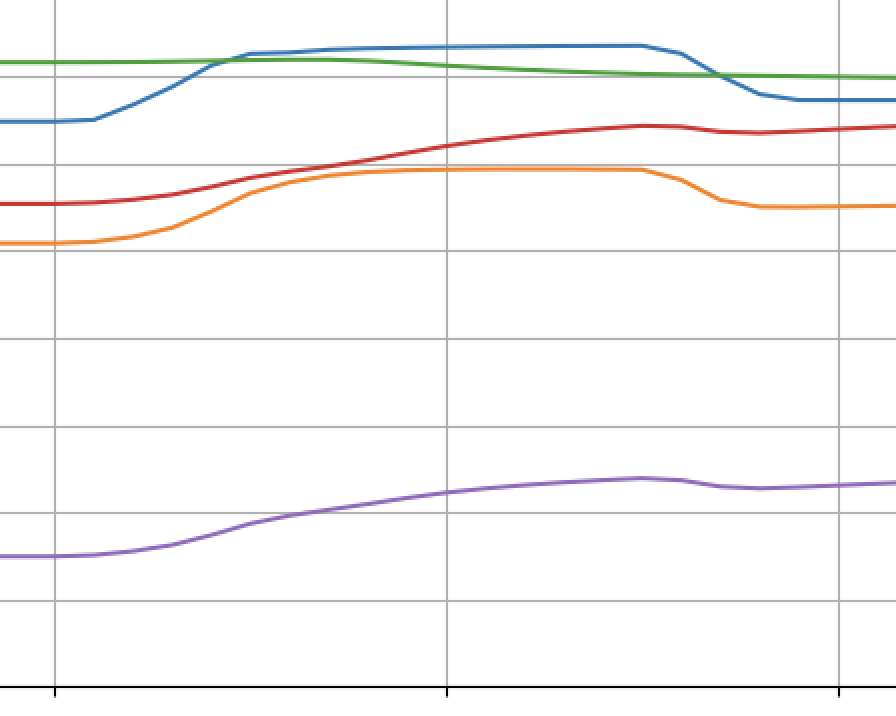}
  \caption{/ku/}
\end{subfigure}

\caption{Vertical displacement trajectories of flesh-point for nine CV syllables. 
Columns: different vowel contexts: /Ca/, /Ci/, /Cu/; Rows: different intitial 
consonant: /pV/, /tV/, /kV/; blue: tongue dorsum; orange: tongue tip; green: upper lips;
red: lower lips; magenta: lower jaw. Time interval is t = 40 msec to t = 160 msec (c.f. Fig. 2)} 
\label{fig:fleshpoint_cv}
\end{figure}

The lips--jaw and tongue--jaw articulatory tradeoff can be most clearly observed by examining the
vertical components of flesh-point trajectories obtained for the upper and lower lips, the tongue tip,
the tongue dorsum, and the lower jaw across different consonants and vowel contexts (see
Fig.~\ref{fig:fleshpoints_midsagittal}, Fig.~\ref{fig:fleshpoint_pa_ti} and Fig.~\ref{fig:fleshpoint_cv}). In
Fig.~\ref{fig:fleshpoint_pa_ti}, the corresponding control-parameter trajectories are additionally
displayed, reflecting the underlying sensorimotor expectations (task-space trajectories) that drive
the articulatory movements.

The example syllable /pa/ (Fig.~\ref{fig:fleshpoint_pa_ti}a) illustrates the following: 

(i) In the upper part of Fig.~\ref{fig:fleshpoint_pa_ti}a (vertical displacement of flesh points) we can see that the 
lower jaw position at the end of the syllable is lower than the vocalic neutral position at the beginning of the 
syllable (i.e., before the onset of consonantal closure formation). This reflects vocalic articulation, because the end of 
the syllable corresponds to the vocalic /a/ posture. During consonant production, however, a high lower-jaw position is 
reached. This highlights the contribution of lower-jaw elevation to the of the lower lip during the production of the 
consonantal closure for /p/.

(ii) The lower part of Fig.~\ref{fig:fleshpoint_pa_ti}a (control parameters; task space paraemters) demonstrates 
the full independence of the consonantal and vocalic control parameters during syllable production: The lip closing 
sensorimotor demand does not interact with sensorimotor demands for vocalic height or vocalic fronting for later vowel 
production. Thus, the gesture interaction as well as gestural articulator tradeoff is not initiated at this level. 

(iii) In the upper part of Fig.~\ref{fig:fleshpoint_pa_ti}a, the trajectories of vertical displacement for the upper and lower 
lip flesh points show that the labial closure is entirely determined by the vocalic control parameter ``lip closing.'' Labial 
closure is visible here as the convergence of the displacement trajectories of the upper and lower lip. The higher position of 
the lower-lip flesh point relative to the upper-lip flesh point—observable as the lip closure progresses—reflects the increasing 
strength of the labial closure during its temporal span.

Comparable findings are already discussed in studies on saturation effects in speech motor control: Once an articulator comes 
into firm contact with another articulator or with a vocal-tract wall, further increases in muscle activation may produce 
little or no additional change in closure (becasue foll closure is full closure) and thus in the resulting acoustic output
related to vocal tract closure \citep{Perkell1997_SpeechMotorControl,Perkell2000_TheorySpeechMotorControl,
Perkell2004_SaturationEffect_Sibilants}. Moreover, this type of articulatory saturation provides a concrete biomechanical basis 
for the quantal relationships between articulation and acoustics described by 
Stevens \citep{Stevens1972_QuantalNature,Stevens1989_OnQuantal}.

(iv) Back to our syllable example /pa/ generated by DYNARTmo (Fig.~\ref{fig:fleshpoint_pa_ti}a): The flesh-point 
displacements of the tongue tip and tongue dorsum reflect not only the vocalic articulation but also the 
co-movement of the tongue with the lower jaw during consonantal closure formation (upward movement). Thus, tongue tip (and front 
part of tongue dorsum) are eleveated during consonant production as well but, with during ongoing time, 
the tongue (tongue tip and tongue dorsum flesh points) already begins to move downwards toward the vocalic /a/ target during 
the consonantal closure interval.

Comparable effects can be seen in the case of Fig.~\ref{fig:fleshpoint_pa_ti}b (syllable /ti/).

(i) The lower-jaw flesh-point displacement increases during consonant production, but not as strongly as in the case of /ta/, 
because tongue-tip elevation is supported not only by an elevation of the lower jaw but also by a co-elevation of the tongue 
dorsum.

(ii) In addition, we observe the upward co-movement of the lower lip, which is caused by the upward movement of the lower jaw 
during consonant production.

(iii) The downward movement of the tongue-dorsum flesh point reflects the coarticulatory behavior of the tongue dorsum during 
the upward movement of the tongue tip that forms the apical closure. During this time interval, the spatial region between 
tongue dorsum and tongue tip exhibits a concave shape.

Figure~\ref{fig:fleshpoint_cv} shows the variation of lower-jaw displacement for CV syllables (with C = plosive) across 
different places of articulation, corresponding to different primary closure-forming articulators (labial, apical, dorsal), 
and in combination with the edge vowels /a/, /i/, and /u/. The resulting 9 C–V combinations yield 9 syllables, which already 
exhibit a large variety of lips–lower-jaw and tongue–lower-jaw tradeoffs.

While Fig.~\ref{fig:fleshpoint_pa_ti} illustrates the complete temporal course of syllable production for two of these 
9 CV syllables, the nine subpanels of Fig.~\ref{fig:fleshpoint_cv} display only the critical temporal interval: the  
articulatory transitions leading into the consonantal closure, the interval of consonantal clsoure, and the time interval of the 
release, i.e., the transition into the subsequent vowel (time interval from t = 40 msec to t = 160 msec).

This figure shows the following:

(i) A strong increase in lower-jaw displacement for both labial and dorsal consonants; in the case of apical consonants, part 
of the articulatory tradeoff between the primary articulator and additional assisting articulators is carried not only by the 
lower jaw but also by the tongue dorsum.

(ii) The saturation effect occurs for bilabial closure in all vowel contexts. This highlights the primacy of consonantal 
articulation during the interval of closure or near-closure formation.

(iii) The characteristic co-movements of articulators that are not involved in forming the consonantal closure 
(e.g., the lips in the case of apical or dorsal closure, or the tongue tip and tongue dorsum in the case of labial closure) 
are clearly recognizable in all panels. These co-movements result solely from the active upward movement of the lower jaw as 
part of the articulatory tradeoff required for forming the consonantal closure.

\section{Discussion and Conclusions}

DYNARTmo provides a simplified but effective framework for simulating articulatory movement patterns and 
articulator synergies. Despite not implementing a full task-dynamic architecture, the model naturally 
reproduces tongue--jaw tradeoffs during closure or constriction formation, reflecting widely reported 
empirical patterns. Through the examples presented---ranging from simple CV sequences to more complex 
clusters---we illustrate how vowel context, consonantal target, and articulator coupling jointly determine 
the resulting movement patterns.

DYNARTmo thus represents a useful tool for visualizing, teaching, and analyzing articulatory synergy and 
constraints, offering an accessible complement to more elaborate task-dynamic models.

\section{Supplementary Material}

The simulation model can be downloaded as a Web App titled \textit{Speech Articulation Trainer} 
(\url{https://speech-articulation-trainer.web.app/}). In order to generate and visualize trajectories 
as exemplified in Fig.~2, start the app and click the buttons: \texttt{VAL}, \texttt{set}, and 
\texttt{+fleshPts}. Then click the buttons \texttt{SND} and \texttt{move}, and finally press the 
\texttt{play}-button. 

The syllable /\textipa{pa}/ will then be simulated, and its video animation in midsagittal view will be 
displayed (unless you have selected a different syllable; /\textipa{pa}/ is the default, already activated 
when the app is started). The flesh-point and control-parameter trajectories will be saved automatically on 
your device in the \texttt{/downloads/} folder as files named 
\texttt{control\_123456789.csv} and \texttt{displacement\_123456789.csv}. 

The Python code for visualizing the data contained in both files is available for download in the folder 
\texttt{py\_code/plot\_displ\_and\_control.py}.

\bibliographystyle{apalike}
\bibliography{references}

\end{document}